\documentclass{article}

\usepackage{PRIMEarxiv}

\usepackage[utf8]{inputenc} 
\usepackage[T1]{fontenc}    

\usepackage{hyperref,url}            
\usepackage{booktabs,multirow}       
\usepackage{amssymb,amsfonts,amsmath}       
\usepackage{nicefrac}   
\usepackage{enumitem}
\usepackage{microtype}      
\usepackage{fancyhdr} 
\usepackage{graphicx,float}  

\usepackage[authoryear]{natbib}

\usepackage[table]{xcolor}

\usepackage{amsmath,amsfonts,bm}

\def\eqref#1{equation~\ref{#1}}

\def\1{\bm{1}}

\DeclareMathAlphabet{\mathsfit}{\encodingdefault}{\sfdefault}{m}{sl}
\SetMathAlphabet{\mathsfit}{bold}{\encodingdefault}{\sfdefault}{bx}{n}

\title{\texttt{SPINET}: Sheaf Protein Inverse Folding Network}

\author{
  Jens Lundsgaard*\textsuperscript{1,2,3}
  \qquad Colin Mikulski*\textsuperscript{4}
  \qquad Zhixuan Yan\textsuperscript{4}
  \qquad Dhananjay Bhaskar\textsuperscript{3--7}\\[0.6em]
  \textsuperscript{1}Department of Computer Sciences \quad
  \textsuperscript{2}Department of Mathematics \\
  \textsuperscript{3}Department of Biomedical Engineering \quad
  \textsuperscript{4}Biophysics Graduate Program \quad
  \textsuperscript{5}Data Science Institute \\
  \textsuperscript{6}Center for Genomic Science Innovation \quad
  \textsuperscript{7}Wisconsin Institute for Translational Neuroengineering\\
  University of Wisconsin--Madison, Madison, WI, USA \\
  *Equal Contribution \qquad Correspondence: \texttt{dhananjay.bhaskar@wisc.edu}
}

\begin{document}

\maketitle
\vspace{-.8cm}

\begin{abstract}
Proteins change shape as they function, yet most inverse folding models predict amino acid sequences from a single, fixed backbone. A central challenge in protein engineering is to design proteins that undergo specific motions, which requires accounting for how their structures change over time. This motivates inverse protein folding conditioned on protein motion. We introduce SPINET, which predicts sequences from molecular dynamics trajectories. It uses cellular sheaves to represent residue interactions within each frame and recurrent units to integrate information across frames, then predicts all amino acids in a single pass. We evaluate SPINET on mdCATH and ATLAS, where it outperforms all evaluated static and ensemble baselines in sequence recovery. On mdCATH, it achieves 56.7\% top-1 recovery, compared with 44.5\% for the strongest static baseline and 40.7\% for the strongest ensemble baseline. We also evaluate whether the predicted sequences are compatible with conformations sampled along the target trajectory. On mdCATH, they achieve a median TM-score of 0.760, and structural recovery favors target conformations over unrelated decoys for 99.5\% of test domains.
\end{abstract}

\keywords{sheaf neural networks \and inverse folding \and molecular dynamics}

\section{Introduction}

A protein's amino acid sequence shapes its three-dimensional structure, which in turn influences its function. Inverse protein folding (IPF) asks which sequence is compatible with a given target backbone. Yet proteins are not rigid: they move between conformational states during catalysis, transport, signaling, and allostery, so a single structure may miss motions relevant to their function \citep{agarwalPowerPitfallsAlphaFold2024, boehrRoleDynamicConformational2009, henzler-wildmanDynamicPersonalitiesProteins2007, stankProteinBindingPocket2016}. As protein design begins to address more complex sequence-structure-function relationships \citep{listovOpportunitiesChallengesDesign2024}, a further goal is to identify sequences compatible with prescribed conformational changes \citep{guoDeepLearningGuided2025}. For example, one could modify the observed motion of a wild-type protein and seek sequence variants intended to follow the new path.

Most IPF models, however, predict sequences from a single static structure \citep{gaoPiFoldEffectiveEfficient2023,10.1093/bioinformatics/btaf666,baiMaskpriorguidedDenoisingDiffusion2025}, even though proteins occupy conformational landscapes in which residue
contacts, orientations, and local environments change over time \citep{10.1093/nar/gkad1084}. These dynamics can provide sequence constraints that are not visible in a single snapshot. More recent approaches incorporate multiple discrete conformational states per
protein \citep{ICLR2026_f94cfd15,yiAllatomInverseProtein2025},
reflecting the fact that a functional protein exists as a conformational ensemble
\citep{yabukarskiEnsemblefunctionRelationshipsQualitative2025}. However, discrete ensembles do not capture the evolving local interactions and intermediate states along a molecular dynamics trajectory. To our knowledge, no IPF model has directly taken such a trajectory as input.

To address this gap, we introduce \texttt{SPINET}, the \textbf{S}heaf \textbf{P}rotein \textbf{I}nverse \textbf{F}olding \textbf{Net}work, an architecture that predicts sequences directly from molecular dynamics trajectories by interleaving spatial and temporal aggregation. For the spatial component, \texttt{SPINET} uses \emph{cellular sheaves} \citep{hansenTowardASpectral2019,bodnarNeuralSheafDiffusion2023a}, which assign a vector space to every node and edge and a learnable restriction map to every incidence. This approach is well suited for proteins, where residues participate in distinct local geometries: each interaction can be mediated by its own learned linear transformation rather than aggregated as a scalar message in a shared space. \texttt{SPINET} pairs this spatial machinery with a recurrent temporal component, applying sheaf message passing at each trajectory frame, integrating frame-wise representations over time, and refining them with a final stack of sheaf message-passing blocks before predicting the amino acid at each position.

We evaluate \texttt{SPINET} on molecular dynamics trajectories from the mdCATH dataset \citep{mirarchi2024mdcath}, with additional sequence-recovery results on ATLAS \citep{10.1093/nar/gkad1084}  reported in the appendix. On mdCATH, \texttt{SPINET} achieves 56.7\% top-1 sequence recovery and a perplexity of 3.74, outperforming both static and ensemble baselines, and structural recovery analysis shows that its predictions are specifically compatible with the target conformational ensemble rather than responsive to arbitrary structural templates. \texttt{SPINET} also achieves state-of-the-art sequence recovery on ATLAS, reaching 51.8\% top-1 recovery and a perplexity of 4.261, compared with 23.1\% recovery and a perplexity of 18.788 for an ablation without learnable sheaf restriction maps.

To summarize, our main contributions are:
\begin{enumerate}[itemsep=1mm,leftmargin=.5cm,labelsep=.5em]
\item A dynamics-aware formulation of inverse protein folding that recovers sequences from protein motion rather than from static structures or discrete ensembles, taking a step toward designing proteins that undergo prescribed conformational changes.
\item A novel sheaf-based architecture that interleaves spatial message passing through learned restriction maps with recurrent units that integrate information across trajectory frames.
\item Evaluation on mdCATH and ATLAS using top-$k$ sequence recovery and perplexity, together with structural recovery on mdCATH using representative trajectory conformations and unrelated decoy templates. We compare against static and ensemble-based baselines and ablations of \texttt{SPINET}.
\item Learned restriction maps and their associated sheaf Laplacians provide representations that can be analyzed and interpreted. In our preliminary analysis, differences in the sheaf-Laplacian spectral gap appear to correspond to differences in protein flexibility.
\end{enumerate}

\section{Related Work}

Most inverse folding methods predict sequences from a single protein backbone. Graph-based approaches represent residues as nodes and their spatial relationships as edges, using geometric features and message
passing to infer which amino acids fit the target structure. GVP-GNN \citep{jingLearningProteinStructure2020} separates scalar features that are invariant to rotation from vector features that transform with the protein's orientation. PiFold
\citep{gaoPiFoldEffectiveEfficient2023} uses backbone distances,
angles, and relative directions to capture both local interactions and global context. ScFold \citep{zhongScFoldGNNbasedModel2025}
reduces residue features over overlapping regions before applying attention, allowing neighboring regions to share information near their boundaries. These architectures differ in how they represent and aggregate geometry, but all condition their predictions on one static structure.

Generative approaches also largely retain the single-backbone
setting. MapDiff \citep{baiMaskpriorguidedDenoisingDiffusion2025}
conditions on the target structure while iteratively denoising an amino acid sequence. It combines geometric graph representations with a masked-residue refinement step to improve its predictions. This changes how a sequence is generated, but the structural input remains a fixed backbone.

DynamicMPNN \citep{ICLR2026_f94cfd15} is a recent method that
conditions autoregressive sequence generation on multiple protein conformations. It encodes distinct structures into a shared representation and generates a sequence for the resulting ensemble. It thus accounts for multiple target states, but treats them as discrete conformations rather than using a molecular dynamics trajectory as input.

IF-MD \citep{brotzakisDesignProteinSequences2025a} addresses the
design of proteins with specified kinetic properties using
molecular dynamics. It uses ProteinMPNN
\citep{dauparasRobustDeepLearning2022a} to assess sequences against structures sampled by molecular dynamics, reweights the simulated ensemble to estimate kinetic properties, and applies Bayesian optimization to propose new sequences. Its objective is therefore to optimize sequences for desired kinetics rather than to recover a sequence directly from an input trajectory, so it is not a directly comparable inverse folding baseline.

In comparison, \texttt{SPINET} conditions sequence recovery on molecular dynamics trajectories, integrating information across their frames. It predicts amino acid identities at all residue positions in a single pass, without autoregressive decoding.

\begin{figure}[H]
    \centering
    \includegraphics[width=0.99\linewidth]{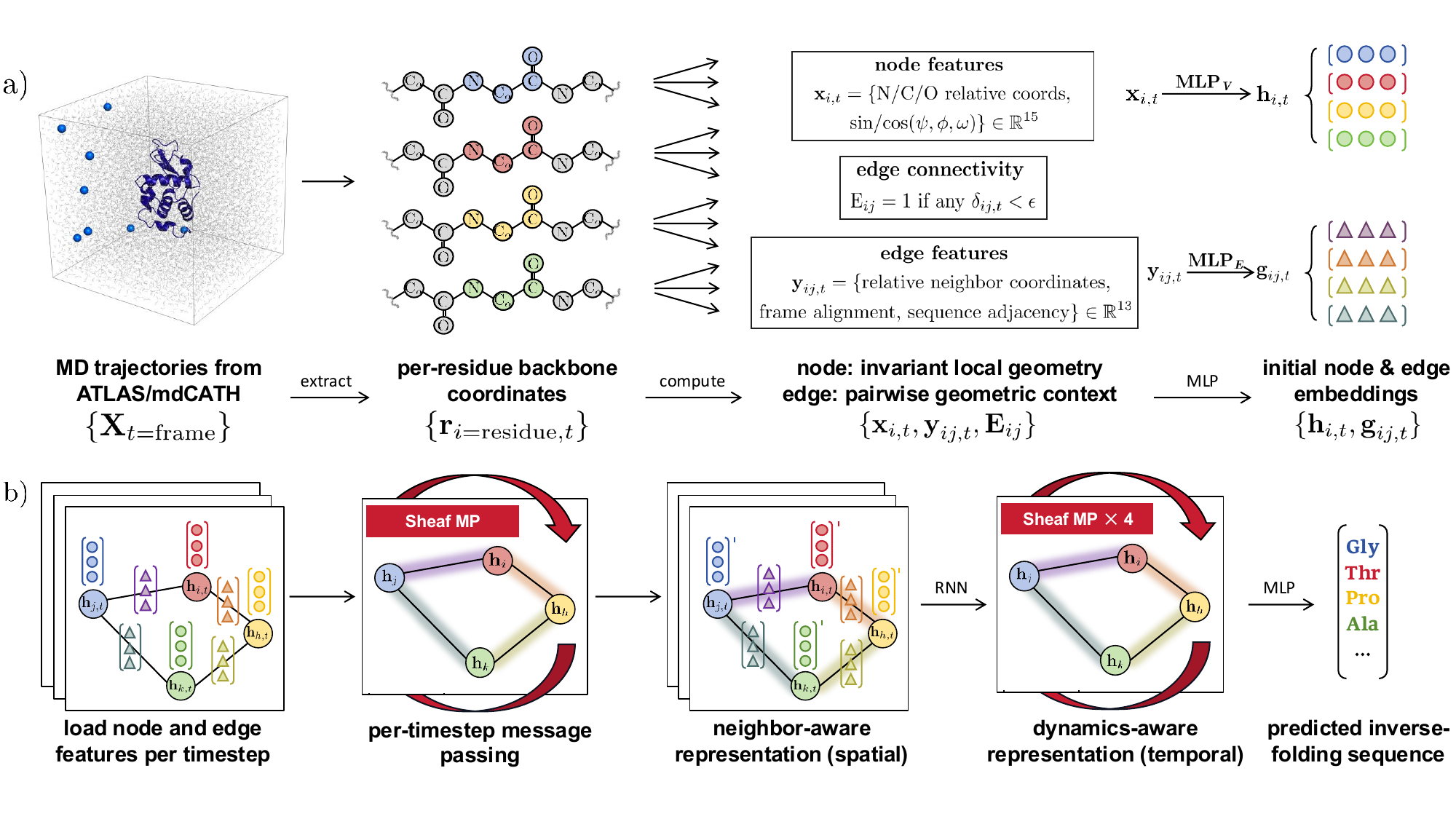}
    \caption{\texttt{SPINET} architecture. (a) Backbone coordinates from a molecular dynamics trajectory are converted into residue-level node features and geometric edge features. (b) Sheaf message passing encodes residue interactions within each frame, an RNN aggregates these representations over time, and additional sheaf message-passing blocks refine them before amino acid classification.}
    \label{fig:SPINET_architecture}
\end{figure}

\section{Background}

A cellular sheaf $\mathcal{F}$ on a graph $G=(V,E)$ assigns a finite-dimensional real vector space $\mathcal{F}(i)$ to each node $i$ and a vector space $\mathcal{F}(e_{ij})$ to each edge $e_{ij}$. It also assigns a linear restriction map $\mathcal{F}_{i \leq e_{ij}}: \mathcal{F}(i) \to \mathcal{F}(e_{ij})$ to each node-edge incidence $i \leq e_{ij}$. In our model, every vector space has dimension $d$, so each restriction map is represented by a matrix in $\mathbb{R}^{d\times d}$.

For a graph with $n$ nodes and stalk dimension $d$, the sheaf Laplacian $\mathcal{L}_{\mathcal{F}}$ is an $nd\times nd$ block matrix \citep{hansenTowardASpectral2019}. Its diagonal blocks are
\[
(\mathcal{L}_{\mathcal{F}})_{ii}
=
\sum_{j\sim i}
\mathcal{F}_{i\leq e_{ij}}^\top
\mathcal{F}_{i\leq e_{ij}},
\]
and its off-diagonal blocks are
\[
(\mathcal{L}_{\mathcal{F}})_{ij}
=
\begin{cases}
-\mathcal{F}_{i\leq e_{ij}}^\top
 \mathcal{F}_{j\leq e_{ij}}, & e_{ij}\in E,\\
\mathbf{0}, & e_{ij}\notin E.
\end{cases}
\]
When $d=1$ and every restriction map is the identity, the sheaf Laplacian reduces to the standard graph Laplacian $\mathcal{L}_G=D-A$.

\section{Methods}

As shown in Fig.~\ref{fig:SPINET_architecture}, \texttt{SPINET} applies sheaf message passing to each trajectory frame, aggregates the resulting representations over time with an RNN, and applies further sheaf message passing before residue classification. We describe each stage below.

\subsection{Node and Edge Features}

Following \citet{gaoPiFoldEffectiveEfficient2023}, we define a local atomic frame at each residue using the orthonormal basis
\[
R =
\left[
\frac{u-v}{\|u-v\|},\,
\frac{u\times v}{\|u\times v\|},\,
\frac{u-v}{\|u-v\|}
\times
\frac{u\times v}{\|u\times v\|}
\right],
\]
where
\[
u=C_\alpha-C,
\qquad
v=N-C_\alpha.
\]
We express the position of atom $A$ in node $i$'s local frame as
\[x_A = R_i^T(A' - C_{\alpha_i})\]
where $R_i$ is the atomic frame of node $i$, $A'_j$ is the raw position of atom $A$, and $C_{\alpha_i}$ is the raw position of node $i$'s $C_\alpha$ atom.

The node features $x_{i,t}$ contain the sine and cosine of the dihedral angles $\phi$, $\psi$, and $\omega$, along with the positions of backbone atoms $C$, $O$, and $N$ in the local atomic frame at time $t$. The edge features $x_{ij,t}$ contain the position of node $j$'s $C_\alpha$ atom in node $i$'s frame, the column-wise cosine similarities between the two atomic frames at time $t$, and a time-independent binary indicator distinguishing covalent bonds from proximity edges.

We add an edge between nodes $i$ and $j$ when $\min_{1\leq t \leq T} \|C_{\alpha_{i,t}} - C_{\alpha_{j,t}} \| < \varepsilon$, for a chosen threshold $\varepsilon > 0$. If this produces more than $k$ edges for node $i$, we retain the $k$ neighbors with the smallest minimum distances to $i$.

\begin{figure}[H]
\centering
\includegraphics[width=0.99\linewidth]{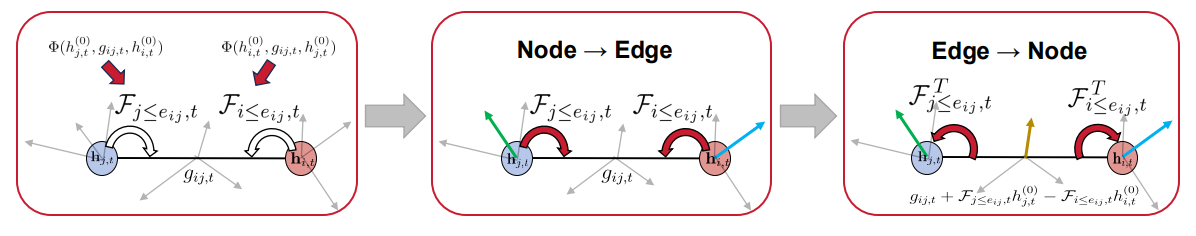}
\caption{\texttt{SPINET} learns a restriction map $\mathcal{F}_{j \le e_{ij}}$ for each incident node--edge pair. During sheaf message passing, the map transforms the neighbor representation $h_j$ into the edge space as $\mathcal{F}_{j\leq e_{ij}}h_j$, where it is combined with the learned edge representation $g_{ij}$. The resulting message is mapped back to the node space and used to update the residue representation $h_i$; the updated representation is then concatenated with the previous $h_i$.}
\label{fig:sheafMP}
\end{figure}

\subsection{Learnable Restriction Maps}

Following \citet{bodnarNeuralSheafDiffusion2023a}, we learn restriction maps using
\[
\Phi_{\theta_{\mathcal F}}:
\mathbb{R}^{2f+h}\rightarrow\mathbb{R}^{d\times d},
\]
where $x_i,x_j\in\mathbb{R}^f$ are node features and $x_{ij},x_{ji} \in \mathbb{R}^h$ are the corresponding directed edge features. For each edge $e_{ij}$, the restriction maps are
\[
\mathcal{F}_{i\leq e_{ij}}
=
\Phi_{\theta_{\mathcal F}}(x_i\|x_{ij}\|x_j),
\qquad
\mathcal{F}_{j\leq e_{ij}}
=
\Phi_{\theta_{\mathcal F}}(x_j\|x_{ji}\|x_i),
\]
where $\|$ denotes concatenation. The order of the inputs allows the model to learn a distinct restriction map for each of the two node--edge incidences. Fig.~\ref{fig:sheafMP} illustrates how these maps transform residue representations during message passing. We impose no additional constraints on the learned maps.

\subsection{Spatiotemporal Embedding}

At each time step $t$, the model learns a sheaf $\mathcal{F}_t$ from the node and edge features of that frame. Separate MLPs project these features into hidden representations:
\[
h_{i,t}^{(0)}
=
\operatorname{MLP}_{\theta_1}(x_{i,t}),
\qquad
g_{ij,t}
=
\operatorname{MLP}_{\theta_2}(x_{ij,t}).
\]
Both representations lie in $\mathbb{R}^{cd}$ and are arranged into $c$ channels of stalk dimension $d$. This allows the same restriction map to act independently on each channel.

We then perform sheaf message passing within each frame:
\[
h_{i,t}^{(1)}
=
\sum_{i\sim j}
\mathcal{F}_{i\leq e_{ij},t}^{\top}
\left[
g_{ij,t}
+
\mathcal{F}_{j\leq e_{ij},t}h_{j,t}^{(0)}
-
\mathcal{F}_{i\leq e_{ij},t}h_{i,t}^{(0)}
\right].
\]
For each neighbor $j$, the restriction maps place the representations of $i$ and $j$ in the shared edge space. The message combines their difference with the learned edge representation $g_{ij,t}$; the transpose of the restriction map then returns it to the space of node $i$. Summing these messages gives a frame-specific representation of the residue's local interactions. The $d\times d$ restriction maps act independently on each channel. In the ablation without sheaves, the frame-wise aggregation becomes
\[
h_{i,t}^{(1)}
=
\sum_{i\sim j}
\left(g_{ij,t}+h_{j,t}^{(0)}\right).
\]
An MLP then acts on the full $cd$-dimensional hidden representation: $$h_{i,t}^{(2)} = \operatorname{MLP}_{\theta_3}(h_{i,t}^{(1)}).$$
For each residue, an RNN processes these frame-wise representations in temporal order. Its final hidden state summarizes the residue's interactions across the trajectory and is passed through an MLP:
\[
z_i
=
\operatorname{MLP}_{\theta_5}
\left(\operatorname{RNN}_{\theta_4}
\left(
\{h_{i,t}^{(2)}\}_{t=1}^{T}
\right)\right).
\]
After temporal aggregation, we reshape $z_i$ into the residue representation $h_i\in\mathbb{R}^{c\times d}$. We learn a new sheaf $\mathcal{F}$ from these temporally aggregated representations, allowing neighboring residues to exchange information about their trajectory-wide behavior. Each subsequent sheaf message-passing block updates $h_i$ as
\[
h_i'
=
\operatorname{MLP}_{\theta_6}
\left(
\sum_{i\sim j}
\mathcal{F}_{i\leq e_{ij}}^{\top}
\left[
\mathcal{F}_{j\leq e_{ij}}(h_jW)
-
\mathcal{F}_{i\leq e_{ij}}(h_iW)
\right]
\right)
+h_i,
\]
where $W\in\mathbb{R}^{d\times d}$ is learned and the restriction maps act independently on each channel. The MLP acts on the flattened representation in $\mathbb{R}^{cd}$; its output is reshaped into $\mathbb{R}^{c\times d}$ before the residual addition.

For the ablation without sheaves, the update becomes
$$
h_i'
=
\operatorname{MLP}_{\theta_6}
\left(
\sum_{i\sim j}h_jW
\right)
+h_i.
$$
We apply $b$ consecutive message-passing blocks.

\subsection{Residue Classification and Loss}

After the final message-passing block, a classification head produces logits for the amino acid classes:
$$
\ell_i
= \operatorname{MLP}_{\theta_5}(h_i).$$
We train the model by minimizing the cross-entropy loss
\[
\mathcal{L}
=
\operatorname{CrossEntropy}
\left(
\{\ell_i\}_{i\in V},
\{y_i\}_{i\in V}
\right),
\]
where $y_i$ is the native amino acid at residue $i$.

\section{Results}

We evaluate \texttt{SPINET} on molecular dynamics trajectories from mdCATH \citep{mirarchi2024mdcath} and ATLAS \citep{10.1093/nar/gkad1084}. Our primary evaluation uses mdCATH, which contains trajectories for 5,398 protein domains. We construct a 70/15/15 split by CATH topology, so that test domains belong to topologies absent from training. We compare \texttt{SPINET} with four static inverse protein folding models - PiFold \citep{gaoPiFoldEffectiveEfficient2023}, GVP-GNN \citep{jingLearningProteinStructure2020}, ScFold \citep{zhongScFoldGNNbasedModel2025}, and MapDiff \citep{baiMaskpriorguidedDenoisingDiffusion2025} - and the ensemble model DynamicMPNN \citep{ICLR2026_f94cfd15}. All methods are trained and evaluated on the same train-validation-test splits. Implementation and baseline-training details are provided in Appendix~\ref{app:implementation-details}. Our code is available on GitHub\footnote{\url{https://github.com/UW-Madison-CBML/SPINET}}.

We first measure how well each model recovers the native amino acid sequence. Top-$k$ recovery is the fraction of residue positions for which the native amino acid appears among the model's $k$ highest-ranked predictions. Perplexity is the exponential of the mean negative log probability assigned to the native amino acids; lower perplexity indicates better calibrated sequence predictions. We then assess whether predicted sequences are compatible with conformations sampled from the target trajectory using AF3 template-based structural recovery.

In Table \ref{tab:mdcath_seqeunce_results}, we report sequence recovery on the 320K mdCATH test set. \texttt{SPINET} achieves 56.7\% top-1 recovery, 89.8\% top-5 recovery, and 97.9\% top-10 recovery, with a perplexity of 3.74. MapDiff, the strongest static baseline by top-1 recovery, reaches 44.5\%, while fine-tuned DynamicMPNN reaches 40.7\%. These comparisons suggest that information across the trajectory helps recover amino acids that are difficult to identify from a single backbone or a pair of conformations. \texttt{SPINET} has 1.5M parameters, fewer than DynamicMPNN, PiFold, ScFold, and MapDiff. 

The ablations separate the contributions of sheaf structure and inter-residue interactions. In the temporal graph neural network (TGNN), we replace sheaf message passing with graph convolutions while retaining recurrent aggregation across trajectory frames. TGNN reaches 31.9\% top-1 recovery, compared with 56.7\% for \texttt{SPINET}. The Traj--MLP baselines remove the graph structure entirely. For each residue, we average its node features over time and use the resulting vector to predict its amino acid independently of the other residues; residue-level predictions are then aggregated to evaluate recovery for each protein. Using all node features, Traj--MLP reaches 40.6\% top-1 recovery. Versions using only dihedral features or only coordinate features reach 28.6\% and 36.9\%, respectively. These results show that time-averaged local geometry contains sequence information, while the full model gains substantially from learning interactions between residues across the trajectory.

\begin{table}[h]
\centering
\caption{Native-sequence recovery on the 320K mdCATH test split. Top-$k$ recovery is the fraction of residues whose native amino acid appears among the $k$ highest-ranked predictions; perplexity measures the probability assigned to native residues. Higher recovery and lower perplexity indicate better performance. Parameter counts are shown for each model.}
\label{tab:mdcath_seqeunce_results}
\setlength{\tabcolsep}{8pt}
\renewcommand{\arraystretch}{1.1}
\resizebox{\linewidth}{!}{
\begin{tabular}{llccccc}
\toprule
Paradigm & Model & Params & top-1 $\uparrow$ & top-5 $\uparrow$ & top-10 $\uparrow$ & Perplexity $\downarrow$ \\
\midrule

\multirow{5}{*}{Dynamic}
& \texttt{SPINET}
& 1.5M
& \textbf{0.567 $\pm$ 0.060}
& \textbf{0.898 $\pm$ 0.038}
& \textbf{0.979 $\pm$ 0.017}
& \textbf{3.736 $\pm$ 0.676} \\

& TGNN
& 1.4M
& 0.319 $\pm$ 0.054
& 0.712 $\pm$ 0.058
& 0.898 $\pm$ 0.039
& 8.872 $\pm$ 1.592 \\

& Traj--MLP (full)
& 23.1K
& 0.406 $\pm$ 0.063
& 0.774 $\pm$ 0.053
& 0.935 $\pm$ 0.031
& 6.301 $\pm$ 1.001 \\

& Traj--MLP (dihedrals)
& 22.5K
& 0.286 $\pm$ 0.059
& 0.643 $\pm$ 0.062
& 0.860 $\pm$ 0.046
& 9.848 $\pm$ 1.438 \\

& Traj--MLP (coords)
& 22.7K
& 0.369 $\pm$ 0.060
& 0.724 $\pm$ 0.052
& 0.904 $\pm$ 0.036
& 7.297 $\pm$ 1.063 \\

\midrule

\multirow{2}{*}{Ensemble}
& DynamicMPNN (re-trained)
& 4.2M
& 0.376 $\pm$ 0.063
& 0.754 $\pm$ 0.054
& 0.919 $\pm$ 0.034
& 7.242 $\pm$ 1.228 \\

& DynamicMPNN (fine-tuned)
& 4.2M
& 0.407 $\pm$ 0.064
& 0.780 $\pm$ 0.051
& 0.929 $\pm$ 0.031
& 6.597 $\pm$ 1.154 \\

\midrule

\multirow{4}{*}{Static}
& PiFold
& 6.6M
& 0.343 $\pm$ 0.086
& 0.710 $\pm$ 0.086
& 0.896 $\pm$ 0.052
& 8.310 $\pm$ 2.444 \\

& GVP
& 928.6K
& 0.270 $\pm$ 0.069
& 0.639 $\pm$ 0.077
& 0.849 $\pm$ 0.054
& 12.473 $\pm$ 3.429 \\

& ScFold
& 7.2M
& 0.396 $\pm$ 0.134
& 0.741 $\pm$ 0.110
& 0.903 $\pm$ 0.060
& 10.520 $\pm$ 5.876 \\

& MapDiff
& 14.7M
& 0.445 $\pm$ 0.085
& 0.835 $\pm$ 0.061
& 0.956 $\pm$ 0.028
& 5.779 $\pm$ 1.665 \\

\bottomrule
\end{tabular}}
\end{table}

Sequence recovery measures agreement with native amino acid labels, but does not establish whether a predicted sequence is compatible with the conformations in the target trajectory. We therefore evaluate structural recovery using the AF3 template-pullback procedure adapted from \citet{ICLR2026_f94cfd15}. For each test domain, we represent trajectory frames with the invariant local-geometric features used by \texttt{SPINET}, apply PCA and $k$-means clustering, and select the observed frame nearest each cluster centroid. This produces five representative conformations, $X_1,\ldots,X_5$, selected independently of the predicted sequences and held fixed across methods. For \texttt{SPINET}, we sample five window-conditioned sequence predictions uniformly along the trajectory.

Given a predicted sequence $S$, we provide each representative conformation $X_j$ to AF3 as a structural template and compare the resulting structure $\hat X_{j,r}$ with $X_j$, where $r$ indexes five AF3 diffusion samples. TM-score and C$_\alpha$ RMSD assess global structural agreement after optimal Kabsch superposition of corresponding C$_\alpha$ atoms, while lDDT assesses preservation of local pairwise distances. For each metric $m$, we average over the five diffusion samples and five representative conformations:
$$ \bar m(S) = \frac{1}{5}\sum_{j=1}^{5}
\frac{1}{5}\sum_{r=1}^{5}
m(\hat X_{j,r},X_j)$$
Table~\ref{tab:structural_recovery} reports the median of these scores across the 788 domains evaluable for every method. Higher TM-score and lDDT, and lower RMSD, indicate stronger recovery.

\texttt{SPINET} achieves a median TM-score of 0.760 and RMSD of 2.98~\AA. These values are close to those of DynamicMPNN (0.756 and 3.11~\AA) and exceed those of the static baselines in global structural recovery. DynamicMPNN has slightly higher lDDT than \texttt{SPINET} (0.789 versus 0.777), indicating better local-distance preservation by this measure. The architectural ablations show larger differences: the TGNN model reaches a TM-score of 0.619 and RMSD of 4.58~\AA, while Traj--MLP reaches 0.209 and 13.23~\AA. Thus, the local geometry that supports residue recovery in Traj--MLP does not, by itself, produce comparably strong structural recovery.

AF3 might also respond to a supplied template even when that template is unrelated to the predicted sequence. To assess target specificity, we evaluate each sequence with one structurally unrelated decoy per domain, chosen to have TM-score $<0.4$ against all five representative target conformations. We compare recovery under target and decoy templates using
$$
\mathrm{TM}_{\mathrm{target/decoy}}
=
\frac{\mathrm{TM}_{\mathrm{target}}}
     {\mathrm{TM}_{\mathrm{decoy}}},
\qquad
\mathrm{RMSD}_{\mathrm{target/decoy}}
=
\frac{\mathrm{RMSD}_{\mathrm{target}}}
     {\mathrm{RMSD}_{\mathrm{decoy}}}.
$$
A larger TM-score ratio and a smaller RMSD ratio indicate greater preference for the target. \texttt{SPINET} has a median TM-score ratio of 6.91 and RMSD ratio of 0.194; target-template TM-score exceeds decoy-template TM-score for 99.5\% of domains. These values are comparable to those of DynamicMPNN (6.90, 0.197, and 98.4\%, respectively) and close to the native-sequence control. The structural recovery therefore favors trajectory-derived conformations over unrelated decoys.

Finally, because \texttt{SPINET} generates window-conditioned predictions, different regions of the same trajectory can generate different candidate sequences. Therefore, we examine the capacity of this candidate set. We additionally evaluate a post-hoc best-of-five upper bound, selecting from five uniformly sampled window predictions the sequence with the highest mean TM-score across the representative conformations. This increases median TM-score from $0.760$ to $0.810$ and reduces RMSD from $2.976$ to $2.360$~\AA, approaching the native benchmark and exceeding it on RMSD and decoy-normalized TM-score. Although this oracle selection is not a direct comparison with single-sequence baselines, it shows that the candidate set frequently contains a highly structure-compatible sequence. This motivates a learned cross-window aggregation or selection mechanism as a natural extension of the current architecture that has strong potential.

\begin{table}[H]
\centering
\caption{Structural recovery and target specificity on the 320K mdCATH test set. TM-score, lDDT, and C$_\alpha$ RMSD compare AF3 predictions with representative target conformations. Target/decoy ratios compare recovery under target and unrelated decoy templates; ``target favored'' is the percentage of domains with a higher TM-score under the target template. Values are medians across the 788 domains evaluable for every method. Higher TM-score, lDDT, and target/decoy TM-score and pLDDT ratios, together with lower RMSD and target/decoy RMSD ratio, indicate better performance. Interquartile ranges and additional metrics appear in Appendix~\ref{app:mdCATH_struct}.}
\label{tab:structural_recovery}

\setlength{\tabcolsep}{5.5pt}
\renewcommand{\arraystretch}{1.08}

\resizebox{\linewidth}{!}{
\begin{tabular}{lccccccc}
\toprule
Sequence source
& TM-score $\uparrow$
& lDDT $\uparrow$
& RMSD (\AA) $\downarrow$
& TM$_{\text{tar/dec}}$ $\uparrow$
& RMSD$_{\text{tar/dec}}$ $\downarrow$
& pLDDT$_{\text{tar/dec}}$ $\uparrow$
& Target favored (\%) $\uparrow$ \\
\midrule

\rowcolor{gray!20} Native
& 0.824
& 0.835
& 2.405
& 7.244
& 0.153
& 1.44
& 99.0 \\

\texttt{SPINET}
& \textbf{0.760}
& 0.777
& \textbf{2.976}
& \textbf{6.912}
& \textbf{0.194}
& \textbf{1.36}
& \textbf{99.5} \\

\rowcolor{gray!20} \texttt{SPINET} (Bo5)$^\dagger$
& 0.810
& 0.804
& 2.360
& 7.370
& 0.154
& 1.42
& 99.6 \\

TGNN
& 0.619
& 0.693
& 4.576
& 6.041
& 0.286
& 1.16
& 99.1 \\

Traj--MLP (full)
& 0.209
& 0.438
& 13.231
& 2.324
& 0.772
& 1.03
& 91.1 \\

\midrule

MapDiff
& 0.729
& 0.765
& 3.323
& 6.473
& 0.219
& 1.18
& 97.7 \\

PiFold
& 0.708
& 0.752
& 3.570
& 6.257
& 0.229
& 1.16
& 97.0 \\

GVP
& 0.651
& 0.714
& 4.127
& 6.026
& 0.260
& 1.17
& 96.1 \\

ScFold
& 0.715
& 0.753
& 3.579
& 6.496
& 0.222
& 1.33
& 95.3 \\

DynamicMPNN
& 0.756
& \textbf{0.789}
& 3.107
& 6.899
& 0.197
& 1.10
& 98.4 \\

\bottomrule
\end{tabular}
}

\vspace{2pt}
\footnotesize
$^\dagger$Post-hoc upper bound obtained by selecting, for each domain, the \texttt{SPINET} candidate with the highest mean TM-score across the five representative conformations.
\end{table}

Full structural-recovery distributions and additional target--decoy metrics are reported in Appendix~\ref{app:mdCATH_struct}. In Appendix~\ref{app:aa_f1} we report recovery by amino acid class, while Appendix~\ref{app:temp-comparison} compares sequence recovery across mdCATH simulation temperatures using an earlier model variant. Finally, we evaluate on the ATLAS dataset in Appendix~\ref{app:atlas-dataset}. On the ATLAS test split, \texttt{SPINET} reaches 51.8\% top-1 recovery and a perplexity of 4.26, compared with 41.3\% and 6.47 for fine-tuned DynamicMPNN, the strongest evaluated baseline by top-1 recovery on that split.

\begin{figure}[H]
    \centering
    \includegraphics[width=0.8\linewidth]{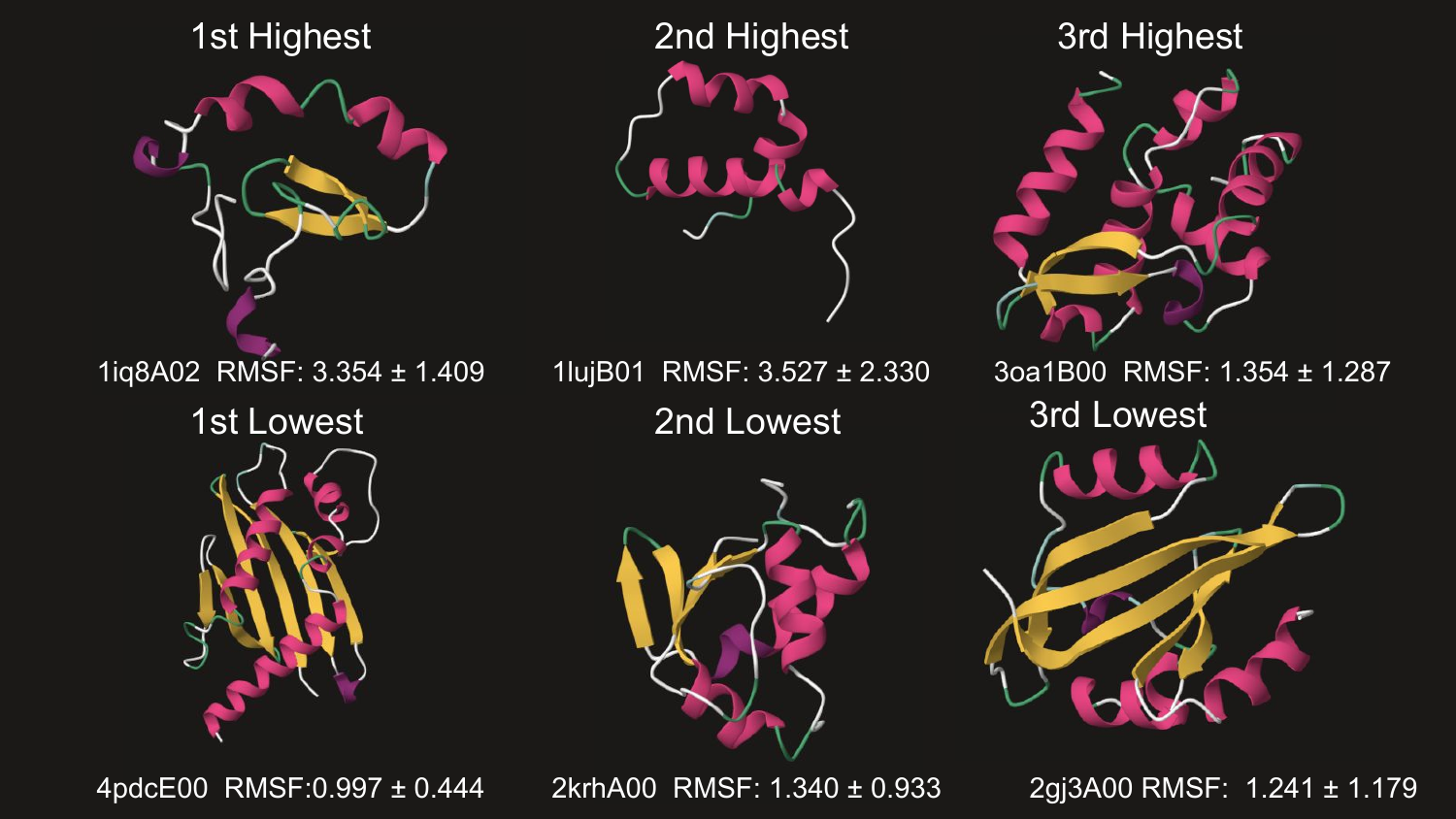}
    \caption{Proteins with the three largest (top row) and three smallest (bottom row) spectral gaps of the learned sheaf Laplacian from the first message-passing block after recurrent aggregation. Each structure is labeled with its protein identifier and root-mean-square fluctuation (RMSF, in \AA) across the trajectory.}
    \label{fig:sheaf_interpretation}
\end{figure}

\subsection{Interpretability of learned restriction maps}

We next examine whether the learned sheaf structure provides information about protein motion. Following the spectral perspective on cellular sheaves \citep{hansenTowardASpectral2019}, we analyze the sheaf Laplacian from the first message-passing block after the recurrent unit. This is the first sheaf block to process representations aggregated across trajectory frames. For each protein, we calculate the difference between the two largest eigenvalues of its learned sheaf Laplacian and compare proteins at the high and low ends of this measure.

Fig.~\ref{fig:sheaf_interpretation} shows the three proteins with the largest and three with the smallest spectral gaps, alongside their root-mean-square fluctuations (RMSF). RMSF is compututed from frames $X_{t=1}^T$ via Kabsch RMSD comparison to the first frame $X_1$. The examples with the largest gaps generally show greater fluctuations than those with the smallest gaps. This qualitative comparison suggests a possible relationship between the learned sheaf spectrum and protein flexibility; establishing that relationship across the full dataset requires further analysis.

\section{Conclusion}

We developed \texttt{SPINET} for inverse protein folding conditioned on molecular dynamics trajectories. By combining sheaf message passing within each frame with recurrent aggregation across frames, the model uses changing residue interactions to predict an amino acid sequence. \texttt{SPINET} outperformed static and ensemble-based baselines in sequence recovery on mdCATH and ATLAS. On the 320~K mdCATH test set, it achieved 56.7\% top-1 recovery, compared with 44.5\% for the strongest static baseline and 40.7\% for the strongest ensemble baseline. Its predicted sequences recovered representative target conformations with a median TM-score of 0.760, and target-template recovery exceeded decoy-template recovery for 99.5\% of evaluable domains in the structural analysis. These results show that protein motion provides useful constraints for recovering amino acid sequences compatible with the conformations sampled during that motion. In protein engineering, our approach could help redesign the motion path of an existing protein and propose sequence variants intended to follow the modified path. More broadly, coupling motion-path generation with \texttt{SPINET} could enable de novo design of proteins for specified conformational changes, including switches, molecular motors, and enzymes that move between functional states.

\paragraph{Limitations.}
Our structural evaluation supplies target conformations to AF3 as templates. It therefore does not establish that the predicted sequences will fold into those conformations or undergo the prescribed motion experimentally. We assessed structural recovery on mdCATH only; our ATLAS evaluation was limited to sequence recovery. We also did not test how performance changes when substantial portions of an input trajectory are withheld or its frames are shuffled. Structural evaluation on ATLAS and tests using shorter, sparsely sampled, or reordered trajectories are important next steps.

Applying \texttt{SPINET} to a new design target also requires a plausible motion path before the intended sequence is known. One possibility is to start with the observed motion of an existing protein, modify that path, and ask which sequence variants could follow it. For a protein designed from scratch, candidate paths could instead be proposed by conformational-sampling models, flow- or diffusion-based backbone generators, coarse-grained simulations, or manually prescribed motions. In either setting, \texttt{SPINET} would serve as the sequence-design component of a larger motion-conditioned pipeline. Testing sequences designed for modified or generated paths, including experimental assessment of their folding, dynamics, and function, remains for future work.

\section*{Acknowledgments}
The authors acknowledge support from the Center for High Throughput Computation at UW--Madison.

\bibliographystyle{plainnat}
\bibliography{iclr2027_conference}  

\clearpage
\newpage

\appendix

\section{Implementation Details}
\label{app:implementation-details}

\subsection{\texttt{SPINET} Hyperparameters and Training}

Unless otherwise stated, \texttt{SPINET} was trained with the hyperparameters listed below. The TGNN and Traj--MLP ablations used the same training hyperparameters wherever applicable.

\begin{tabular}{llcc}
\toprule
Hyperparameter  & Value\\
\midrule
$d$ & 8 \\
$c$ &  16 \\
$b$ &  6 \\
$\varepsilon$ & 7.5 Angstroms \\
$k$ & 32\\
Learning Rate & 0.001 \\
Epochs & 20 \\
Batch Size & $< 3000$ Nodes\\
Time Subsequence Length & 128 \\
Time Subsequence Step  & 32 \\
\bottomrule
\end{tabular}

\subsection{Baseline Model Training}

We trained the static baselines on relaxed RCSB Protein Data Bank structures \citep{bermanProteinDataBank2000,burleyUpdatedResourcesExploring2025} matching the PDB and chain IDs in mdCATH and ATLAS. We aligned each structure's sequence to its corresponding dataset sequence, retained the aligned residues, and used the same training, validation, and test splits as for the dynamic models. PiFold was trained for up to 200 epochs with a batch size of 8 and learning rate $1\times10^{-4}$; training stopped if validation perplexity did not improve for 10 epochs. We report results from its best validation-perplexity checkpoint. ScFold and GVP were each trained for 100 epochs at a learning rate of $1\times10^{-3}$. ScFold used a batch size of 8, while GVP used batches of at most 3,000 nodes.

For MapDiff, we followed the authors' masked-prior pretraining procedure for 200 epochs using Adam, a batch size of 8, and a peak learning rate of $5\times10^{-4}$. We then trained the denoising model for up to 200 epochs and selected epoch 34 based on mean validation perplexity on mdCATH. Because of its high computational expense, we trained and evaluated MapDiff only on mdCATH.

We trained DynamicMPNN from scratch for up to 200 epochs and selected the checkpoint with the lowest mean validation perplexity (epoch 89 on mdCATH and epoch 87 on ATLAS). We used a learning rate of $1\times10^{-3}$ and an effective batch size of 32, obtained through two gradient-accumulation steps of 16. DynamicMPNN was originally trained on pairs of chains from one or two PDB entries selected for substantial structural dissimilarity. To construct conformer pairs from our MD trajectories, we coarsely sampled frames at intervals of 4 for mdCATH and 25 for ATLAS, then computed a pairwise TM-score matrix. We selected a pool of 10 frames with the highest pairwise TM-scores. During training, we randomly selected two conformers from this pool and applied Gaussian noise; during validation and testing, we selected the pair with the highest pairwise TM-score, following the authors' procedure.

We also fine-tuned published DynamicMPNN weights trained on single chains with two conformational states. The fine-tuning setup used the same conformer-selection procedure and effective batch size of 32. We selected the checkpoint with the lowest mean validation perplexity from at most 200 epochs: epoch 196 on mdCATH and epoch 197 on ATLAS.

We supplied two MD-derived conformers per protein to reflect DynamicMPNN's intended two-state setting. As discussed by \citet{ICLR2026_f94cfd15}, frames from an MD trajectory may span a limited range of conformations, so adding more frames could provide nearly identical structural states. The resulting MD-derived pairs may nevertheless be less distinct than the conformational states for which DynamicMPNN was designed.

\section{Structural Recovery and Target-Decoy Specificity on mdCATH}
\label{app:mdCATH_struct}

Table~\ref{tab:af3_pullback_320} provides the full structural-recovery results, including interquartile ranges (IQR) and analyses restricted to conformations for which the native sequence achieves a TM-score above $0.5$ or $0.7$. On the full set, \texttt{SPINET} reaches a median TM-score of 0.760, while its post-hoc best-of-five selection reaches 0.810. Recovery improves across methods when the analysis is restricted to conformations that AF3 recovers more accurately for the native sequence.

Table~\ref{tab:decoy_structural_recovery} reports structural recovery when AF3 is supplied with unrelated decoy templates. Decoy TM-scores are low and RMSDs are high across the evaluated sequences, indicating that template conditioning alone generally does not recover the decoy structures.

Table~\ref{tab:target_decoy_specificity} compares recovery under target and decoy templates for the same domains. For \texttt{SPINET}, the median target-to-decoy TM-score ratio is 6.912, and target-template TM-score exceeds decoy-template TM-score for 99.5\% of domains. The best-of-five selection increases the ratio to 7.370, while Traj--MLP shows substantially weaker target specificity.

\begin{table}[H]
\centering
\caption{AF3 template-based structural recovery on the 320K mdCATH test set. The all-states analysis reports domain-level median (IQR) over the 788 evaluable domains for every method. Sensitivity analyses retain representative conformations for which the native sequence achieves a TM-score above $0.5$ or $0.7$. Best-of-five (Bo5) selects the \texttt{SPINET} candidate with the highest mean TM-score within each subset. Higher TM-score, lDDT, and pLDDT and lower RMSD indicate better recovery.}
\label{tab:af3_pullback_320}
\setlength{\tabcolsep}{5pt}
\renewcommand{\arraystretch}{1.1}
\resizebox{.95\linewidth}{!}{
\begin{tabular}{llcccc}
\toprule
Subset & Sequence source & TM-like $\uparrow$ & lDDT $\uparrow$ & RMSD (\AA) $\downarrow$ & pLDDT $\uparrow$ \\
\midrule
\rowcolor{gray!20}
\multirow{10}{*}{All states}
 & Native
& 0.824 (0.615--0.922)
& 0.835 (0.751--0.887)
& 2.405 (1.486--4.529)
& 83.30 (77.08--87.71) \\

& $\texttt{SPINET}$
& 0.760 (0.540--0.871)	
& 0.777 (0.688--0.818)	
& 2.976 (2.054--5.457)	
& 73.17 (69.17--76.82) \\

& \texttt{SPINET} (Bo5)
& \textbf{0.810 (0.643--0.901)}
& \textbf{0.804 (0.735--0.841)}
& \textbf{2.360 (1.736--3.892)}
& \textbf{73.97 (69.41--77.91)} \\

& TGNN
& 0.619 (0.362--0.779)	
& 0.693 (0.589--0.744)	
& 4.576 (3.023--8.059)	
& 67.93 (63.78--73.05) \\

& Traj--MLP (full)
& 0.209 (0.105--0.478)
& 0.438 (0.358--0.578)
& 13.231 (8.425--17.190)
& 61.44 (55.00--66.85) \\

& MapDiff
& 0.729 (0.457--0.861)	
& 0.765 (0.667--0.820)	
& 3.323 (2.170--5.759)	
& 76.81 (71.34--81.48) \\

& PiFold
& 0.708 (0.430--0.844)
& 0.752 (0.646--0.809)
& 3.570 (2.331--6.573)
& 76.57 (70.90--81.05) \\

& GVP 
& 0.651 (0.336--0.817)	
& 0.714 (0.590--0.772)	
& 4.127 (2.603--8.115)	
& 71.62 (66.67--75.89) \\

& ScFold 
& 0.715 (0.413--0.856)	
& 0.753 (0.642--0.811)	
& 3.579 (2.249--6.817)	
& 75.34 (69.36--80.23) \\

& DynamicMPNN 
& 0.756 (0.503--0.875)
& 0.789 (0.696--0.833)
& 3.107 (2.031--5.654)
& 77.57 (72.63--81.85) \\

\midrule
\rowcolor{gray!20}
\multirow{10}{*}{Native TM $>0.5$}
&  Native
& 0.878 (0.767--0.942)
& 0.862 (0.813--0.900)
& 1.834 (1.232--2.950)
& 84.84 (79.72--88.75) \\

& $\texttt{SPINET}$
& 0.811 (0.689--0.890)	
& 0.794 (0.746--0.829)	
& 2.501 (1.863--3.709)	
& 74.02 (70.75--77.61) \\

& \texttt{SPINET} (Bo5)
& \textbf{0.835 (0.734--0.906)}
& \textbf{0.814 (0.771--0.845)}
& \textbf{2.142 (1.647--3.000)}
& \textbf{74.65 (70.98--78.41)} \\

& TGNN
& 0.692 (0.532--0.804)	
& 0.713 (0.651--0.754)	
& 3.845 (2.732--5.760)	
& 67.78 (63.77--72.51) \\

& Traj--MLP (full)
& 0.282 (0.134--0.565)
& 0.476 (0.377--0.617)
& 11.931 (7.239--15.989)
& 60.32 (53.68--66.08) \\

& MapDiff
& 0.797 (0.653--0.888)	
& 0.791 (0.724--0.836)	
& 2.705 (1.894--4.104)	
& 77.78 (72.87--82.35) \\

& PiFold
& 0.778 (0.620--0.876)
& 0.780 (0.713--0.826)
& 2.902 (2.045--4.483)
& 77.33 (71.94--81.62) \\

& GVP 
& 0.744 (0.557--0.847)	
& 0.741 (0.669--0.787)	
& 3.250 (2.261--5.346)	
& 72.09 (67.11--76.39) \\

& ScFold 
& 0.781 (0.627--0.881)	
& 0.778 (0.714--0.827)	
& 2.879 (1.979--4.522)	
& 76.55 (70.99--80.96) \\

& DynamicMPNN 
& 0.818 (0.685--0.897)
& 0.807 (0.756--0.845)
& 2.499 (1.787--3.901)
& 78.53 (74.45--82.68) \\

\midrule
\rowcolor{gray!20}
\multirow{10}{*}{Native TM $>0.7$}
&  Native
& 0.903 (0.829--0.950)
& 0.875 (0.836--0.906)
& 1.615 (1.152--2.318)
& 85.60 (80.82--89.08) \\

& $\texttt{SPINET}$
& 0.840 (0.755--0.900)	
& 0.804 (0.766--0.834)	
& 2.260 (1.784--3.067)	
& 74.25 (71.16--77.81) \\

& \texttt{SPINET} (Bo5)
& \textbf{0.857 (0.793--0.913)}
& \textbf{0.824 (0.790--0.850)}
& \textbf{1.953 (1.569--2.541)}
& \textbf{75.09 (71.72--78.92)} \\

& TGNN
& 0.731 (0.605--0.821)	
& 0.721 (0.671--0.758)	
& 3.477 (2.588--5.059)	
& 67.46 (63.76--71.76) \\

& Traj--MLP (full)
& 0.333 (0.145--0.610)
& 0.493 (0.384--0.635)
& 11.429 (6.657--15.800)
& 59.57 (53.04--65.53) \\

& MapDiff
& 0.829 (0.723--0.899)	
& 0.803 (0.751--0.842)	
& 2.420 (1.790--3.545)	
& 78.20 (73.40--82.63)\\

& PiFold
& 0.811 (0.709--0.888)
& 0.793 (0.739--0.833)
& 2.597 (1.935--3.805)
& 77.61 (72.38--81.89) \\

& GVP 
& 0.781 (0.640--0.864)	
& 0.753 (0.694--0.794)	
& 2.907 (2.130--4.437)	
& 72.24 (67.60--76.30) \\

& ScFold 
& 0.818 (0.704--0.892)	
& 0.792 (0.740--0.834)	
& 2.551 (1.849--3.818)	
& 76.87 (71.76--81.19) \\

& DynamicMPNN 
& 0.845 (0.751--0.906)
& 0.817 (0.773--0.850)
& 2.251 (1.689--3.286)
& 78.87 (74.87--82.97) \\

\bottomrule
\end{tabular}}
\end{table}

\begin{table}[H]
\centering
\caption{AF3 structural recovery using templates unrelated to the target conformations. Values are domain-level median (IQR); the number of evaluable domains is shown for each method. Low TM-scores and high RMSDs indicate weak recovery of the decoy structures.}
\label{tab:decoy_structural_recovery}

\setlength{\tabcolsep}{6pt}
\renewcommand{\arraystretch}{1.08}

\resizebox{.95\linewidth}{!}{
\begin{tabular}{lccccc}
\toprule
Sequence source
& Decoy TM-like $\uparrow$
& Decoy lDDT $\uparrow$
& Decoy RMSD (\AA) $\downarrow$
& Decoy pLDDT $\uparrow$
& $n$ domains \\
\midrule

\rowcolor{gray!20} Native
& 0.104 (0.085--0.125)
& 0.302 (0.245--0.364)
& 16.157 (13.565--19.030)
& 55.04 (42.16--72.48)
& 796 \\

\texttt{SPINET}
& 0.106 (0.089--0.122)
& 0.301 (0.245--0.364)
& 15.990 (13.547--18.838)
& 52.22 (41.08--63.39)
& 796 \\

TGNN
& 0.102 (0.084--0.118)
& 0.297 (0.244--0.356)
& 16.731 (14.031--19.558)
& 56.61 (44.12--67.05)
& 796 \\

Traj--MLP (full)
& 0.091 (0.072--0.109)
& 0.274 (0.233--0.333)
& 18.621 (16.117--21.673)
& 53.90 (43.87--63.31)
& 796 \\

\midrule

MapDiff
& 0.105 (0.084--0.126)
& 0.303 (0.248--0.362)
& 16.029 (13.524--19.448)
& 62.18 (47.76--72.84)
& 788 \\

PiFold
& 0.105 (0.086--0.125)
& 0.306 (0.248--0.365)
& 16.056 (13.471--19.297)
& 61.70 (46.78--73.64)
& 788 \\

GVP
& 0.101 (0.079--0.121)
& 0.299 (0.243--0.364)
& 16.532 (13.877--19.809)
& 57.43 (43.73--69.76)
& 788 \\

ScFold
& 0.103 (0.084--0.122)
& 0.299 (0.240--0.363)
& 16.422 (13.830--19.854)
& 53.56 (42.51--66.36)
& 788 \\

DynamicMPNN
& 0.102 (0.084--0.121)
& 0.304 (0.246--0.360)
& 16.157 (13.538--19.136)
& 65.20 (51.09--76.18)
& 796 \\

\bottomrule
\end{tabular}
}
\end{table}

\begin{table}[H]
\centering
\caption{Target-decoy structural specificity on the 320K mdCATH test set. Values are domain-level median (IQR) over the 788 domains evaluable for every method. Ratios compare recovery under target and decoy templates; ``favored'' gives the percentage of domains with better recovery under the target template. Higher TM-like, lDDT, and pLDDT ratios and lower RMSD ratios indicate greater target specificity.}
\label{tab:target_decoy_specificity}

\setlength{\tabcolsep}{4.5pt}
\renewcommand{\arraystretch}{1.08}

\resizebox{.95\linewidth}{!}{
\begin{tabular}{lccccccccc}
\toprule
\multirow{2}{*}{Sequence source}
& \multicolumn{4}{c}{Target/decoy ratio}
& \multicolumn{4}{c}{Favored (\%)}
& \multirow{2}{*}{$n$ domains} \\
\cmidrule(lr){2-5} \cmidrule(lr){6-9}
& TM $\uparrow$
& lDDT $\uparrow$
& RMSD $\downarrow$
& pLDDT $\uparrow$
& TM
& lDDT
& RMSD
& pLDDT
& \\
\midrule

\rowcolor{gray!20} Native
& 7.244 (5.586--8.900)
& 2.654 (2.035--3.475)
& 0.153 (0.085--0.313)
& 1.44 (1.05--1.94)
& 99.0
& 98.5
& 98.5
& 87.1
& 788 \\

\texttt{SPINET}
& 6.912 (5.275--8.271)
& 2.439 (1.875--3.206)
& 0.194 (0.119--0.370)
& 1.36 (1.13--1.78)
& 99.5
& 98.6
& 98.9
& 95.7
& 788 \\

\texttt{SPINET} (Bo5)
& \textbf{7.370 (5.706--8.865)}
& \textbf{2.595 (1.964--3.326)}
& \textbf{0.154 (0.097--0.282)}
& \textbf{1.42 (1.14--1.83)}
& \textbf{99.6}
& \textbf{98.9}
& \textbf{99.6}
& \textbf{94.5}
& 788 \\

TGNN
& 6.041 (3.836--7.391)
& 2.201 (1.670--2.871)
& 0.286 (0.175--0.536)
& 1.16 (1.04--1.49)
& 99.1
& 97.0
& 98.2
& 92.4
& 788 \\

Traj--MLP (full)
& 2.324 (1.329--4.969)
& 1.471 (1.154--2.274)
& 0.772 (0.455--0.921)
& 1.03 (1.00--1.28)
& 91.1
& 87.2
& 86.4
& 78.7
& 788 \\

\midrule

MapDiff
& 6.473 (4.487--8.049)
& 2.412 (1.816--3.100)
& 0.219 (0.128--0.408)
& 1.18 (1.02--1.54)
& 97.7
& 97.1
& 97.3
& 88.3
& 788 \\

PiFold
& 6.257 (4.067--7.897)
& 2.336 (1.741--3.052)
& 0.229 (0.137--0.478)
& 1.16 (1.02--1.56)
& 97.0
& 96.7
& 96.2
& 87.7
& 788 \\

GVP
& 6.026 (3.628--7.637)
& 2.261 (1.635--3.006)
& 0.260 (0.148--0.536)
& 1.17 (1.01--1.59)
& 96.1
& 95.8
& 95.1
& 87.6
& 788 \\

ScFold
& 6.496 (4.040--8.131)
& 2.358 (1.723--3.202)
& 0.222 (0.130--0.475)
& 1.33 (1.03--1.77)
& 95.3
& 94.9
& 94.8
& 89.5
& 788 \\

DynamicMPNN
& 6.899 (4.982--8.375)
& 2.501 (1.918--3.211)
& 0.197 (0.117--0.401)
& 1.10 (1.01--1.48)
& 98.4
& 98.5
& 98.2
& 84.6
& 788 \\

\bottomrule
\end{tabular}
}
\end{table}

\section{Sequence Recovery by Amino Acid Class}
\label{app:aa_f1}

\begin{figure}[H]
    \centering
    \includegraphics[width=1.0\linewidth]{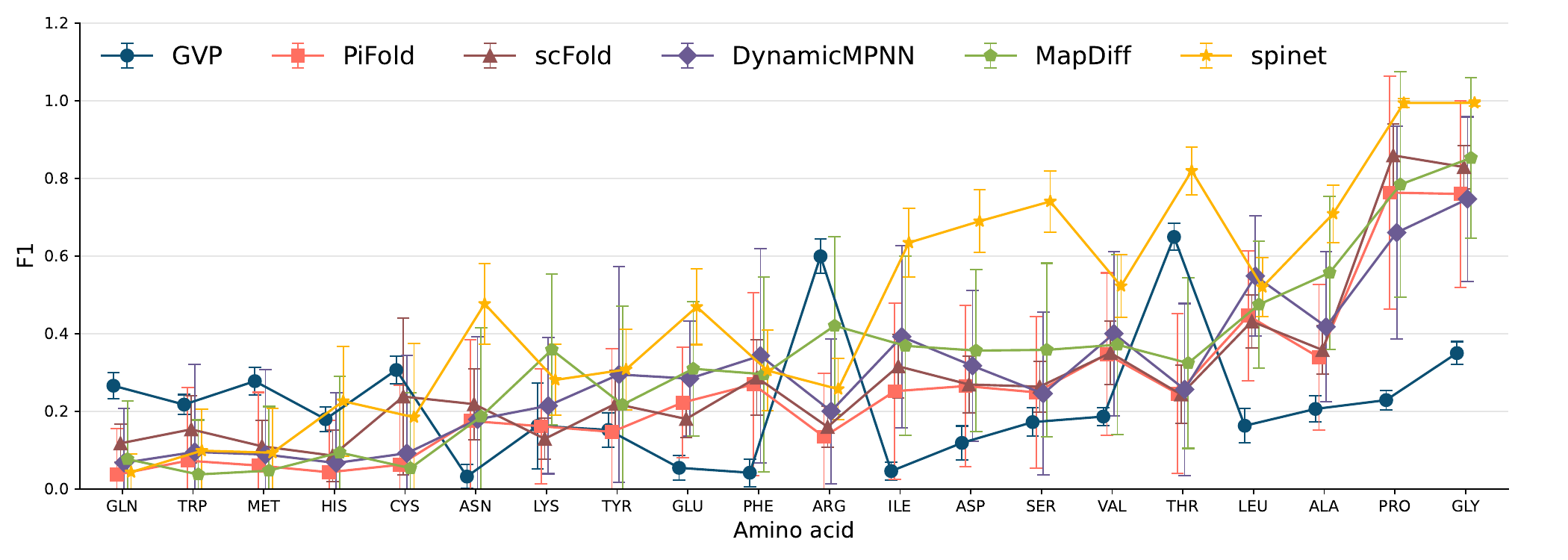}
    \caption{Per-residue average F1 scores across models, evaluated on the test split, with standard deviation error bars.}
    \label{fig:f1_scores}
\end{figure}

Fig.~\ref{fig:f1_scores} shows mean F1 scores by amino acid class for \texttt{SPINET} and the baseline models on the mdCATH test set; error bars indicate standard deviations. \texttt{SPINET} recovers glycine and proline particularly well, consistent with their distinctive backbone geometry. GVP shows relatively strong recovery of arginine and threonine.

\section{Sequence Recovery Across mdCATH Simulation Temperatures}
\label{app:temp-comparison}

Table~\ref{tab:mdCATH-temp} reports \texttt{SPINET}'s sequence recovery and perplexity across the simulation temperatures available in mdCATH. Top-1 recovery increases from 56.4\% at 320~K to 70.0\% at 450~K, while perplexity decreases from 3.712 to 2.356. This trend is consistent with higher-temperature trajectories exposing additional sequence-relevant conformational variation, although the comparison does not establish that variation as the cause of the improvement.

Note that these experiments used an earlier \texttt{SPINET} variant because of overfitting in the temperature-comparison runs. Relative to the main model, it omitted the MLP after the RNN and dropout before the RNN, and used $c=8$ rather than $c=16$. All other hyperparameters were unchanged, and each model was trained for eight epochs.

\begin{table}[H]
\centering
\caption{Sequence recovery and perplexity across mdCATH simulation temperatures. Higher top-$k$ recovery and lower perplexity indicate better sequence prediction.}
\label{tab:mdCATH-temp}
\setlength{\tabcolsep}{8pt}
\renewcommand{\arraystretch}{1.1}
\resizebox{.7\linewidth}{!}{
\begin{tabular}{lccccc}
\toprule
Temperature  & top-1 $\uparrow$ & top-5 $\uparrow$ & top-10 $\uparrow$ & Perplexity $\downarrow$\\
\midrule
450K &   \textbf{0.700 $\pm$ 0.082} & \textbf{0.958 $\pm$ 0.037} & \textbf{0.993 $\pm$ 0.011} & \textbf{2.356 $\pm$ 0.643} \\
413K &     $0.644 \pm 0.074$ & $0.934 \pm 0.039$ & $0.988 \pm 0.014$ & $2.841 \pm 0.663$\\
379K &     $0.598 \pm 0.068$ & $0.910 \pm 0.039$ & $0.982 \pm 0.016$ & $3.330 \pm 0.672$\\
348K &   $0.572 \pm 0.062$ & $0.896 \pm 0.040$ & $0.979 \pm 0.017$ & $3.613 \pm 0.663$ \\
320K &  $0.564 \pm 0.063$ & $0.890 \pm 0.042$ & $0.976 \pm 0.019$ & $3.712 \pm 0.671$ \\
\bottomrule
\end{tabular}}
\end{table}

\section{Sequence Recovery on ATLAS}
\label{app:atlas-dataset}

The ATLAS dataset consists of 1,938 protein trajectories. We excluded 18 because of data loading issues, leaving 1,920 proteins for analysis. We randomly partitioned these proteins into five equally sized sets of 384. Three sets formed the training split ($n=1,152$), one formed the validation split ($n=384$), and one formed the test split ($n=384$). Since ATLAS was assembled to represent diverse protein structures \citep{10.1093/nar/gkad1084}, we split the proteins at random rather than grouping them by structural topology, as we did for mdCATH.

Table~\ref{tab:ATLAS-seq} reports sequence recovery on the ATLAS test split. \texttt{SPINET} achieves 51.8\% top-1 recovery and a perplexity of 4.261, outperforming the evaluated baselines. Its top-1 recovery is lower than the 56.7\% reported on mdCATH in Table~\ref{tab:mdcath_seqeunce_results}. The decrease is larger for the TGNN ablation, whose top-1 recovery falls from 31.9\% on mdCATH to 23.1\% on ATLAS. These results suggest that the sheaf-based model remains effective on the smaller ATLAS dataset.

\begin{table}[H]
\centering
\caption{Sequence recovery and perplexity on the ATLAS test split for \texttt{SPINET}, its ablations, and static and ensemble baselines. Higher top-$k$ recovery and lower perplexity indicate better performance.}
\label{tab:ATLAS-seq}
\setlength{\tabcolsep}{8pt}
\renewcommand{\arraystretch}{1.1}
\resizebox{.9\linewidth}{!}{
\begin{tabular}{llccccc}
\toprule
Paradigm & Model & Params & top-1 $\uparrow$ & top-5 $\uparrow$ & top-10 $\uparrow$ & Perplexity $\downarrow$ \\
\midrule

\multirow{5}{*}{Dynamic}
& \texttt{SPINET}
& 1.5M
& \textbf{0.518 $\pm$ 0.051}
& \textbf{0.878 $\pm$ 0.033}
& \textbf{0.972 $\pm$ 0.015}
& \textbf{4.261 $\pm$ 0.667} \\

& TGNN
& 1.4M
& 0.231 $\pm$ 0.069
& 0.591 $\pm$ 0.080
& 0.820 $\pm$ 0.053
& 18.788 $\pm$ 6.644 \\

& Traj--MLP (full)
& 23.1K
& 0.310 $\pm$ 0.053
& 0.693 $\pm$ 0.049
& 0.887 $\pm$ 0.037
& 8.575 $\pm$ 1.126 \\

& Traj--MLP (dihedrals)
& 22.5K
& 0.268 $\pm$ 0.050
& 0.621 $\pm$ 0.052
& 0.844 $\pm$ 0.039
& 10.206 $\pm$ 1.305 \\

& Traj--MLP (coords)
& 22.7K
& 0.276 $\pm$ 0.044
& 0.672 $\pm$ 0.053
& 0.868 $\pm$ 0.038
& 9.751 $\pm$ 1.186 \\

\midrule

\multirow{2}{*}{Ensemble}
& DynamicMPNN (re-trained)
& 4.2M
& $0.338 \pm 0.058$ 
& $0.726 \pm 0.059$ 
& $0.904 \pm 0.033$ 
& $8.272 \pm 1.499$ \\

& DynamicMPNN (fine-tuned)
& 4.2M
& $0.413 \pm 0.064$ 
& $0.787 \pm 0.054$ 
& $0.932 \pm 0.029$ 
& $6.470 \pm 1.340$ \\

\midrule

\multirow{3}{*}{Static}
& PiFold
& 6.6M
& 0.386 $\pm$ 0.067
& 0.763 $\pm$ 0.061
& 0.921 $\pm$ 0.032
& 7.174 $\pm$ 1.815 \\

& GVP
& 928.6K
& $0.386 \pm 0.068$ 
& $0.761 \pm 0.061$ 
& $0.921 \pm 0.033$ 
& $7.248 \pm 1.920$ \\

& ScFold
& 7.2M
& 0.314 $\pm$ 0.087
& 0.670 $\pm$ 0.093
& 0.865 $\pm$ 0.061
& 20.735 $\pm$ 11.925 \\

\bottomrule
\end{tabular}}
\end{table}

\end{document}